\documentclass[11pt, a4paper, logo, copyright, table]{googledeepmind}

\definecolor{navyblue}{HTML}{0071BC}
\definecolor{hotpink}{HTML}{FF0080}

\definecolor{oai-white}{HTML}{FFFFFF}
\definecolor{oai-black}{HTML}{000000}
\definecolor{oai-red}{HTML}{FF4500}
\definecolor{oai-green}{HTML}{51DA4C}
\definecolor{oai-blue}{HTML}{0000FF}
\definecolor{oai-yellow}{HTML}{FFF639}
\definecolor{oai-magenta}{HTML}{FF45FF}
\definecolor{oai-cyan}{HTML}{00FFFF}
\definecolor{oai-orange}{HTML}{FE7600}
\definecolor{oai-violet}{HTML}{8A2BE2}
\definecolor{oai-brown}{HTML}{A0522D}
\definecolor{oai-green-050}{HTML}{F4FFF4}
\definecolor{oai-green-100}{HTML}{E9FFE8}
\definecolor{oai-green-200}{HTML}{D9FFD8}
\definecolor{oai-green-300}{HTML}{C9FFC7}
\definecolor{oai-green-400}{HTML}{A6FFA3}
\definecolor{oai-green-500}{HTML}{7CF178}
\definecolor{oai-green-600}{HTML}{51DA4C}
\definecolor{oai-green-700}{HTML}{3FA93B}
\definecolor{oai-green-800}{HTML}{2D712A}
\definecolor{oai-green-900}{HTML}{193718}
\definecolor{oai-gray-000}{HTML}{FFFFFF}
\definecolor{oai-gray-100}{HTML}{FAFAFA}
\definecolor{oai-gray-200}{HTML}{F5F5F5}
\definecolor{oai-gray-300}{HTML}{E5E5E5}
\definecolor{oai-gray-400}{HTML}{FFB7A4}
\definecolor{oai-gray-500}{HTML}{CDCDCD}
\definecolor{oai-gray-600}{HTML}{A8A8A8}
\definecolor{oai-gray-700}{HTML}{747474}
\definecolor{oai-gray-800}{HTML}{393939}
\definecolor{oai-gray-900}{HTML}{000000}
\definecolor{visual}{HTML}{A50E0E}       
\definecolor{linguistic}{HTML}{174EA6}   
\definecolor{relational}{HTML}{E37400}   
\definecolor{egocentric}{HTML}{0D652D}

\usepackage[authoryear, sort&compress, round]{natbib}
\usepackage{rotating}

\usepackage[utf8]{inputenc} 
\usepackage[T1]{fontenc}    
\usepackage{hyperref}       
\usepackage{url}            
\usepackage{booktabs}       
\usepackage{amsfonts}       
\usepackage{nicefrac}       
\usepackage{microtype}      
\usepackage{xcolor}         
\definecolor{darkblue}{rgb}{0, 0, 0.7}
\hypersetup{colorlinks=true, citecolor=darkblue, linkcolor=darkblue, urlcolor=darkblue}

\usepackage{enumitem}
\usepackage[noend]{algpseudocode}
\usepackage{algorithm}
\usepackage{algorithmicx}
\usepackage{mathtools}
\usepackage{nccmath}
\usepackage{multirow}
\usepackage{bigdelim}
\usepackage{color, colortbl}
\usepackage{adjustbox}
\usepackage{amsthm}
\usepackage{makecell}
\usepackage{times}
\usepackage{latexsym}
\usepackage{tabularx}
\usepackage{array}
\usepackage{multirow}
\usepackage{siunitx}
\usepackage{graphicx} 
\usepackage{enumitem}
\usepackage{xspace}
\usepackage{caption}
\usepackage{wrapfig}
\usepackage{subcaption}

\usepackage{dcolumn}
\newcolumntype{d}{D{.}{.}{-1}}
\newcolumntype{z}{D{(}{\ (}{1.1}}
\usepackage{bbm}
\usepackage[procnames]{listings}

\definecolor{custom_green}{rgb}{0.0, 0.5, 0.0}
\definecolor{custom_red}{rgb}{1.0, 0.01, 0.24}

\usepackage{makecell}

\usepackage[many]{tcolorbox}

\newcommand{\worldwideweb}{\raisebox{-0.2ex}{\includegraphics[height=1.05em]{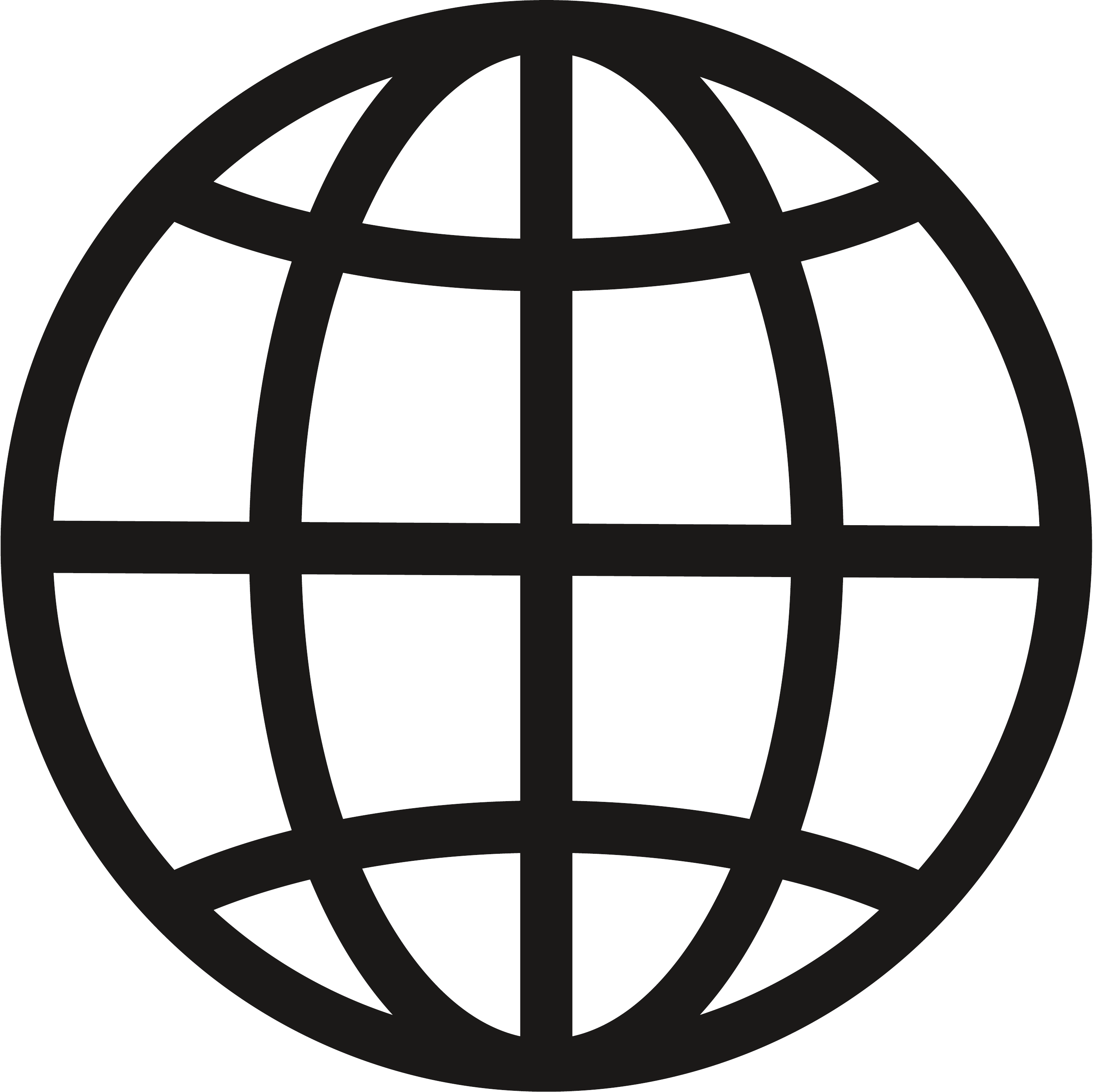}}\xspace}

\newcommand{\github}{\raisebox{-0.2ex}{\includegraphics[height=1.05em]{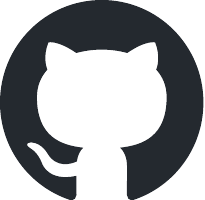}}\xspace}
\newcommand{\huggingface}{\raisebox{-0.2ex}{\includegraphics[height=1.05em]{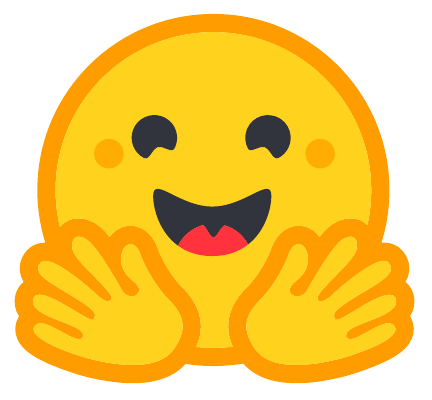}}\xspace}

\title{UGC-VideoCaptioner: An Omni UGC Video Detail Caption Model and New Benchmarks}

\correspondingauthor{my-email@openinterx.com}

\author{
  Peiran Wu$^{1,2*}$,
  Yunze Liu$^{2\dagger}$,
  Zhengdong Zhu$^{2}$,
Enmin Zhou$^{2}$,
  Junxiao Shen$^{1,2}$\\
  $^1$\textbf{University of Bristol}\quad
  $^2$\textbf{Memories.ai Research}\\ 
  $^*$Work done during an internship at Memories.ai Research\\
  $\dagger$Corresponding Author
}

\begin{abstract}
\vspace{-0.5cm}
  {
  \worldwideweb\ \href{https://memories.ai/}{\text{Project Web}} \quad
  \github\ \href{https://github.com/WPR001/UGC_VideoCaptioner}{\text{Evaluation Code}} \quad
   \huggingface\ \href{https://huggingface.co/collections/openinterx/ugc-videocap-6845e290580112a1834737c4}{\text{UGC-VideoCap}}
  }
\vspace{1em}

Real-world user-generated videos, especially on platforms like TikTok, often feature rich and intertwined audio-visual content. However, existing video captioning benchmarks and models remain predominantly visual-centric, overlooking the crucial role of audio in conveying scene dynamics, speaker intent, and narrative context. This lack of full-modality datasets and lightweight, capable models hampers progress in fine-grained, multimodal video understanding.
To address these challenges, we introduce \textbf{UGC-VideoCap}, a new benchmark and model framework specifically designed for detailed, omnimodal captioning of short-form, user-generated videos. Unlike prior datasets, UGC-VideoCap emphasizes balanced integration of audio and visual modalities, featuring 1,000 TikTok videos annotated through a structured three-stage human-in-the-loop pipeline covering audio-only, visual-only, and joint audio-visual semantics. The benchmark also includes 4,000 carefully crafted QA pairs probing both unimodal and cross-modal understanding.
Alongside the dataset, we propose \textbf{UGC-VideoCaptioner-3B}, a 3B-parameter captioning model distilled from Gemini-2.5 Flash. Using a novel two-stage training strategy—supervised fine-tuning followed by Group Relative Policy Optimization (GRPO)—our approach enables efficient adaptation from limited data while maintaining competitive performance. Together, our benchmark and model offer a high-quality foundation and a data-efficient solution for advancing omnimodal video captioning in unconstrained, real-world UGC settings.

\end{abstract}

\begin{document}
\maketitle

\section{Introduction}
\label{sec:introduction}

Generating detailed video captions has long been a key goal in video-language research. Traditional efforts have predominantly focused on describing visual content—such as actions, objects, scenes, and temporal dynamics—within videos~\citep{wang2022git,xu2023mplug,yan2022videococa}, often overlooking the complementary role of audio. This gap is especially pronounced in real-world video domains such as User-Generated Content (UGC) on TikTok and YouTube, and cinematic content where rich audio-visual interplay is central to semantic comprehension. In such settings, audio signals—including speech, music, ambient noise, and sound effects carry critical information that is not visually inferable.

Prior to the advent of Large Language Models (LLMs), earlier approaches were constrained by limited model capacity and weak generalization. Even when trained on large-scale datasets like HowTo100M~\citep{miech2019howto100m} or VideoCC3M~\citep{nagrani2022learning}, models produced only shallow or incomplete descriptions. Audio-based models, such as WaveNet~\citep{oord2016wavenet} and Whisper~\citep{radford2023robust}, remained unimodal, lacking the ability to reason jointly over audio and visual signals. As a result, most existing benchmarks and models are inherently visual-centric and fail to reflect the full multimodal nature of real-world video content.

Recent advances in multimodal LLMs (MLLMs), such as LLaVA-OneVision~\citep{li2024llava}, Qwen2.5-VL~\citep{bai2025qwen2}, and InternVL3~\citep{zhu2025internvl3}, along with post-training strategies like supervised fine-tuning (SFT) and reinforcement learning with human feedback (RLHF), have significantly improved visual understanding capabilities. These models have shown strong performance in captioning~(AuroraCap), inference~(VideoCap-R1~\citep{meng2025videocap}, ST-R1~\citep{wu2025st}), and domain-specific applications~(FMBench~\citep{wu2024fmbench}, MedVLM-R1~\citep{pan2025medvlm}). However, their reliance on visual-only input leaves a major gap: few models genuinely support full omnimodal input. Among open-source systems, only Qwen2.5-Omni~\citep{xu2025qwen2}, MiniCPM-o-2\_6~\citep{yao2024minicpm}, and Gemma3n~\citep{gemma3n} enable joint audio-visual understanding, while closed-source models like the Gemini series from Google also demonstrate such capability. This highlights a fundamental deficiency in current benchmarks and training regimes—there is a lack of large-scale, high-quality omnimodal annotation for real-world videos, especially for short-form UGC content.

\vspace{0.5em}
\noindent\textbf{To address this deficiency, we introduce \textit{UGC-VideoCap}, the first large-scale benchmarks explicitly designed for detailed captioning and QA over short-form, audio-visual UGC videos.} Our key contributions include:

\begin{itemize}
    \item We present an omnimodal benchmark built on 1,000 short TikTok videos (under 1 minute), each enriched with diverse and prominent audio signals. A rigorous three-stage human annotation pipeline is applied: (\textbf{i})~\textit{audio-only annotation} (e.g., number of speakers, voice type, background music, sound effects), with corresponding audio detail captions; (\textbf{ii})~\textit{visual-only annotation} (e.g., OCR text, background transitions, motion dynamics, object types), with corresponding visual captions; and (\textbf{iii})~\textit{audio-visual joint annotation}, yielding a coherent and semantically rich omnimodal caption.
    
    \item We construct a fine-grained QA benchmark with over \textbf{4,000} manually curated open-ended and multiple-choice questions that comprehensively probe both visual and auditory aspects of understanding, providing a challenging and diagnostic evaluation protocol.

    \item We propose a new captioning model, \textbf{UGC-VideoCaptioner (3B)}, trained using a novel two-stage distillation framework. In Stage~1, we utilize Gemini-2.5-Flash to auto-label 20,000 TikTok videos with detailed captions. In Stage~2, we fine-tune using only 2,000 human-aligned captions via SFT, followed by lightweight RLHF. This process achieves performance comparable to full 20k SFT models, demonstrating that high-quality model adaptation can be achieved with significantly reduced human supervision.
\end{itemize}

\noindent We hope our benchmark and training framework will serve as a catalyst for future research in omni video understanding, particularly in real-world domains like UGC and cinema, where rich multimodal cues and fine-grained semantics are essential. The rarity and cost of acquiring fully-aligned omnimodal annotations make this resource especially valuable for benchmarking and training next-generation multimodal LLMs.

\section{Related Work}

\begin{table*}

\centering
\renewcommand{\arraystretch}{1.3}
\resizebox{0.75\textwidth}{!}{\small
\begin{tabular}{l c c c c c}
\toprule
Benchmark & Theme & \#~Videos & \#~QA pairs & Audio in Caption & Visual in Caption \\
\midrule
VATEX~\citep{wang2019vatex} & Human & 4,478 & 4,478 & \ding{55} & \ding{52} \\
\rowcolor{gray!15}
DREAM-1K~\citep{wang2024tarsier} & Open & 1,000 & 6,298 & \ding{55} & \ding{52} \\
\rowcolor{gray!15}
MSR-VTT~\citep{xu2016msr} & Open & 2,990 & 2,990 & \ding{55} & \ding{52} \\
\rowcolor{gray!15}
VDC~\citep{chai2024auroracap} & Open & 1,027 & 96,902 & \ding{55} & \ding{52} \\
VidCapBench~\citep{chen2025vidcapbench} & UGC & 643 & 10,644 & \ding{55} & \ding{52} \\
\textbf{UGC-VideoCap (Ours)} & UGC & 1,000 & 3,975 & \ding{52} & \ding{52} \\
\bottomrule
\end{tabular}
}
\caption{\textbf{Benchmark comparison} for video caption evaluation. UGC-VideoCap has comprehensive and detailed caption evaluation content. Compared with the past video caption benchmarks that mainly focus on the visual part in the annotation process, we propose an omnimodal video caption benchmark.}
\label{tab:bench_comparison}
\end{table*}

\noindent\textbf{Multimodal Large Language Model for Video.}
Video understanding is a critical area within the broader field of vision-language research, aiming to empower models with the ability to interpret and reason about the dynamic visual content presented in videos~\citep{cheng2024videollama,shen2024longvu,wu2025st}. Traditionally, visual content has been parsed through feature extraction using visual encoders~\citep{liu2025map,liu2025videomap}. However, with the rapid advancement of large language models (LLMs), there has been a growing trend toward the development of Multimodal Large Language Models (MLLMs) to support more comprehensive video understanding.
In recent years, major technology companies such as Google, OpenAI, Meta, and Alibaba have made significant progress in this domain, continuously updating their MLLMs—examples include Gemini 2.5~\citep{gemini}, GPT-4o~\citep{hurst2024gpt}, and Qwen2.5-VL~\citep{bai2025qwen2}. These models demonstrate impressive capabilities in processing and reasoning over visual data. Nevertheless, it is widely acknowledged that visual content alone does not capture the full semantic richness of videos, particularly in platforms such as TikTok or Movie, where audio plays an equally crucial role. Therefore, we argue that models relying solely on visual inputs are inherently limited in their ability to perform holistic video understanding.

\noindent\textbf{Video Caption Model and Benchmark.}
The goal of video captioning is to describe the video in terms of several key aspects, thus aiding understanding~\citep{doveh2023dense}, retrieval~\citep{ma2025drvideo} and reasoning~\citep{wu2025st}. As shown in Table~\ref{tab:bench_comparison}, most of the previous video caption benchmarks focus mainly on the visual information in the video, e.g., MSRVTT~\citep{xu2016msr}, VDC~\citep{chai2024auroracap} and VidCapBench~\citep{chen2025vidcapbench}. However, it is well known that there is not only visual information in the video, but also audio information which is crucial for video understanding, especially for UGC, Movie and other types of videos. Often a good audio caption can play a key role in the video understanding task. We can see that there are many excellent omni models such as Gemini~\citep{gemini}, Qwen2.5-omni~\citep{xu2025qwen2}, etc.~\citep{liu2025nexus,zhang2023video}, but a comprehensive omni video detail caption benchmark and small-parameter models are still lacking.
\section{UGC Video Detail Caption Benchmark}
\label{sec:methods}

\subsection{Overview}

\begin{figure}[t!]
    \centering
        \includegraphics[width=0.8\textwidth]{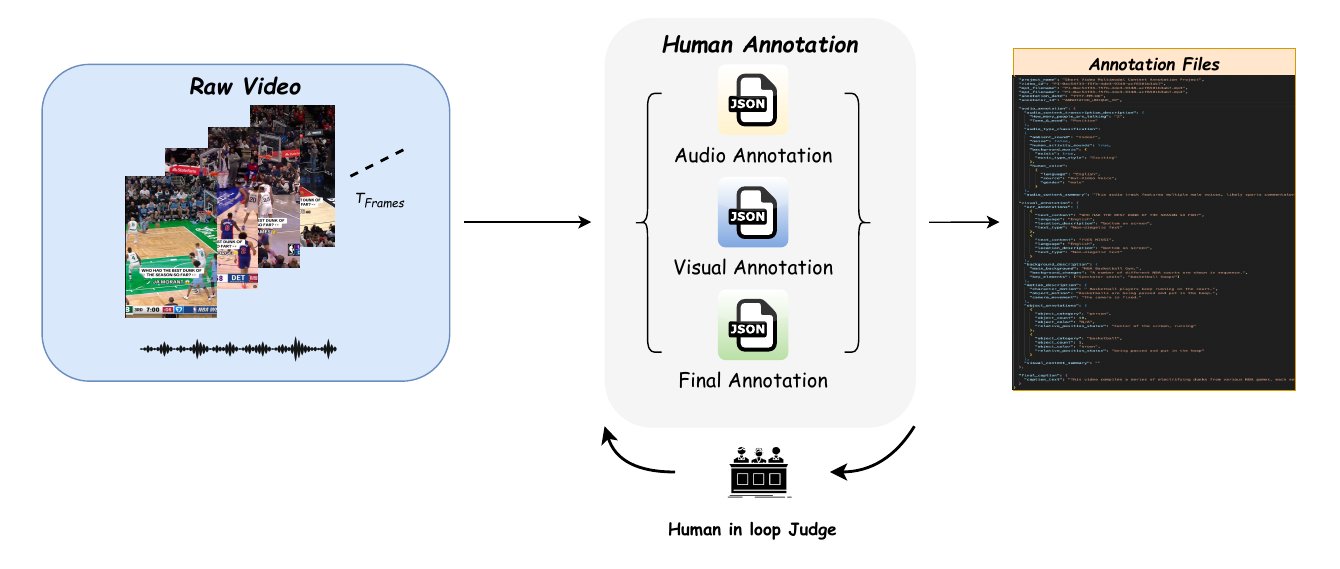}
    \caption{
    Benchmark Collation Pipeline. This pipeline unifies disparate raw video annotation data into a standardised format for consistent processing. QA pairs are then generated from question templates with fixed rules. To ensure quality, manual validation is performed at all key stages, with multiple people cycling through low quality and ambiguous annotated content to assess whether to re-labelling.}
    \label{fig:pipeline}
\end{figure}

We present \textbf{UGC-VideoCap}, a comprehensive benchmark designed to quantitatively evaluate the capability of multimodal large language models (MLLMs) in generating detailed video captions, particularly for short-form videos containing both visual and audio information. UGC-VideoCap includes around 4,000 high-quality question-answer (QA) pairs derived from 1,000 TikTok videos, each with rich vocal tracks and diverse content. 
As illustrated in Figure~\ref{fig:pipeline}, the QA pairs are categorized into three distinct types: \textit{audio-focused}, \textit{visual-focused}, and \textit{comprehensive caption}. In addition to detailed caption annotations, we provide fine-grained attribute labels, including phonetic features (e.g., number of speakers, gender), acoustic conditions (e.g., background music presence and type), and visual properties (e.g., OCR content, object types, background changes). Our evaluation framework is twofold: (1) single-aspect QA evaluation and (2) holistic caption generation. A robust omnimodal model should perform well in both settings, whereas a detail-oriented captioning model may only need to generate accurate final captions. We provided 15-20 audio/visual annotations as raw information, but we only selected some of them that could reflect the characteristics of UGC videos to construct the UGC-VideoCap benchmark.

\subsection{Benchmark Construction}

\noindent\textbf{Data Collection and Standardization.}  
We initiate benchmark construction by curating a diverse set of 1,000 short TikTok videos (each under 60 seconds) that contain meaningful audio segments of at least 5 seconds in length. All videos are manually annotated and subsequently standardized into a unified metadata schema to support structured QA generation. This ensures consistency in annotation quality and enables flexible benchmark expansion in future iterations.

\begin{figure}[t!]
    \centering
        \includegraphics[width=\textwidth]{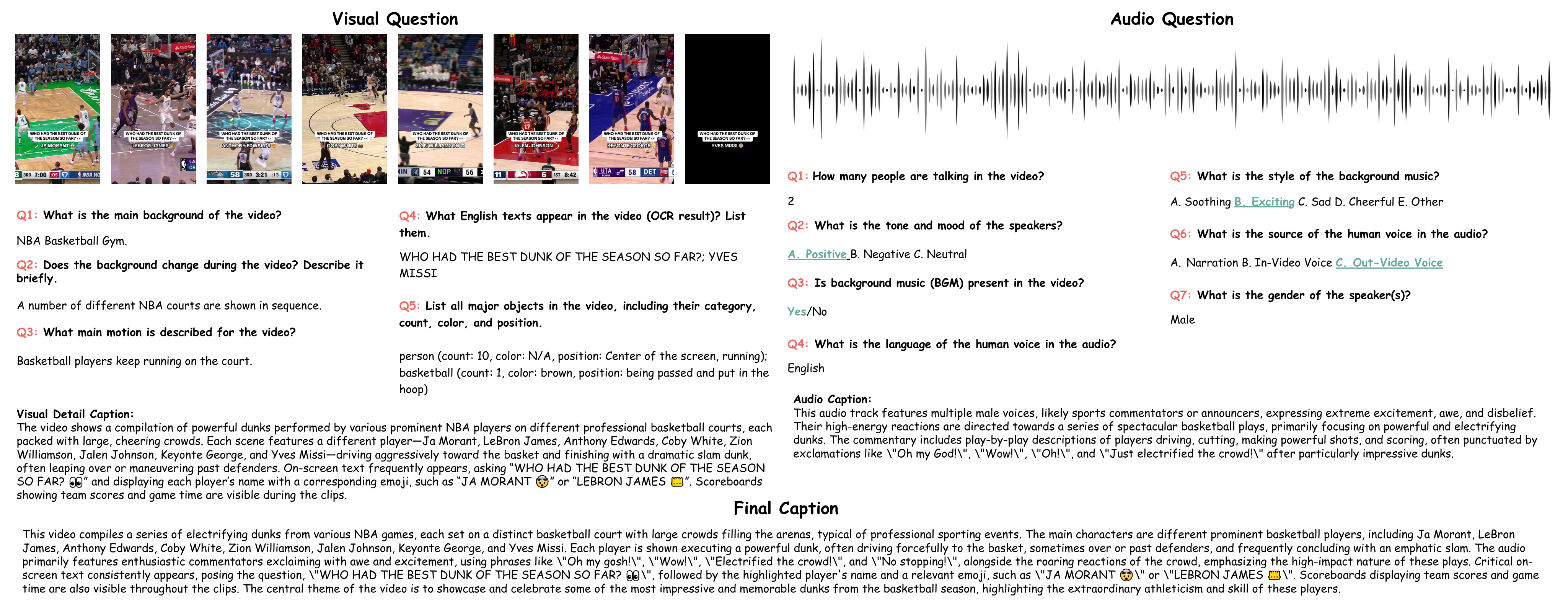}
    \caption{
    Demonstration of tasks for \textbf{UGC-VideoCap} benchmark. It has many meta information for audio and visual. And we select some of them to be our benchmark QA pairs which are visual detail description, audio track detail description and final detail caption.
    }
    \label{fig:sample}
\end{figure}

\noindent\textbf{Question-Answer Generation.}  
The entire data processing and annotation pipeline required over 350 hours of human effort, from initial filtering to final QA generation. We designed a structured QA generation protocol tailored to assess diverse dimensions of video understanding. As shown in Figure~\ref{fig:sample}, we generate three categories of QA pairs:
\begin{itemize}
    \item \textbf{Visual QA}: Focused on scene dynamics, object presence, OCR content, and background changes.
    \item \textbf{Audio QA}: Encompassing speaker attributes (e.g., count, gender), acoustic features (e.g., music genre), and environmental sounds.
    \item \textbf{Comprehensive QA}: Final captioning questions based on both visual and audio modalities, representing a complete semantic understanding of the video.
\end{itemize}
Each video is annotated with a \textit{visual-only caption}, an \textit{audio-only caption}, and a \textit{final omnimodal caption}. These annotations form the basis for both QA and generation tasks. We believe \textbf{UGC-VideoCap} sets a new standard for evaluating omnimodal video understanding. It is among the first benchmarks to emphasize the integral role of audio in video captioning, especially in real-world UGC settings. We posit that full-modality video understanding will become a critical frontier in the next phase of multimodal AI research.

\noindent\textbf{Human-in-the-Loop Annotation Verification.}  
To ensure high annotation fidelity, we adopt a rigorous human-in-the-loop quality assurance protocol. Every batch of 50 annotated video samples is independently reviewed by two expert annotators. Reviewers assess the accuracy, completeness, and clarity of all annotation components—including audio attributes, visual descriptors, and full audiovisual captions.
Errors are defined as factual inaccuracies, omissions of salient content, or inconsistencies with the video. If the error rate in a batch exceeds 3\%, the batch is rejected and returned for re-annotation. Discrepancies between annotators are resolved through arbitration or consensus-based review. This dual-judge protocol ensures consistency and minimizes subjectivity, while establishing a feedback loop between annotators and reviewers. Such a systematic process is critical for maintaining benchmark integrity and provides high-quality supervision for downstream evaluation and training tasks.

\subsection{Evaluation on UGC-VideoCap Benchmark}

\begin{figure*}[t!]
    \captionsetup{type=table}
    \centering
    \fontsize{4.6pt}{4.4pt}\selectfont
    \setlength\tabcolsep{3pt} 
    \renewcommand{\arraystretch}{1.2} 

    \begin{adjustbox}{width=\textwidth}
    \begin{tabular}{r|cc|cc|ccc|cccc}
    & & &
    \rotatebox{75}{Voice Source} &
    \rotatebox{75}{Tone} &
    \rotatebox{75}{OCR.} &
    \rotatebox{75}{Background} &
    \rotatebox{75}{Objects} &
    \rotatebox{75}{Audio} &
    \rotatebox{75}{Visual} &
    \rotatebox{75}{Detail} &
    \rotatebox{75}{Average} \\
    Methods & Rank & Avg. &
    \multicolumn{2}{c}{\cellcolor{orange!10}Audio Detail} &
    \multicolumn{3}{c}{\cellcolor{yellow!10}Visual Detail} &
    \multicolumn{4}{c}{\cellcolor{cyan!10}Final Caption} \\
    \hline
    \rowcolor{navyblue!5}
    \multicolumn{1}{l|}{\textcolor{black}{\textit{Proprietary Models (API)}}} & & & & & & & & & & & \\
    Gemini-2.5-pro & \cellcolor{oai-green-200}{2} & 64.3 & 71.5 & 62.6 & 0.693 & 51.3 & 38.7 & 70.8 & 75.8 & 74.8 & 73.78 \\
    
    Gemini-2.5 Flash & \cellcolor{oai-green-400}{1} & \textbf{67.4} & 78.2 & 64.1 & 0.694 & 53.6 & 43.9 & \textbf{74.2} & \textbf{78.8} & \textbf{77.2} & \textbf{76.73} \\
    
    \hline
    \rowcolor{navyblue!5}
    \multicolumn{1}{l|}{\textcolor{black}{\textit{Open-source Models}}} & & & & & & & & & & & \\
    Gemma-3n-E4B-it & 6 & 38.1 & 65.6 & 46.2 & 0.124 & 4.8 & 15.9 & 54.2 & 54.6 & 51.2 & 53.3 \\
    
    Gemma-3n-E2B-it & 7 & 32.6 & 37.2 & 28.7 & 0.095 & 9.3 & 22.3 & 52.0 & 52.4 & 49.6 & 51.3 \\
    
    Qwen2.5-Omni-7B & \cellcolor{oai-green-200}{3} & 50.9 & 86.6 & 44.9 & 0.644 & 21.6 & 15.7 & \textbf{57.6} & 59.4 & 57.0 & 58.0 \\
    
    Qwen2.5-Omni-3B & 5 & 48.6 & 93.0 & 46.9 & 0.561 & 18.4 & 18.3 & 48.2 & 55.6 & 52.6 & 52.2 \\
    
    MiniCPM-o-2.6-8B & 4 & 48.9 & 80.9 & 29.0 & 0.672 & 26.1 & 9.9 & 46.0 & \textbf{70.4} & \textbf{62.4} & \textbf{59.6} \\
    
    \end{tabular}
    \end{adjustbox}

    \vspace{-0.2cm}
    \caption{\textbf{Evaluation on UGC-VideoCap Benchmark.} We provided a detailed evaluation of the audio and visual, and finally designed a detailed caption with audio and video. Regarding the OCR score we are considering the average of the 4 NLP metrics (BLEU-1, ROUGE-1, ROUGE-2, and ROUGE-L), scaled to percentage for fair comparison in the average calculation.}
    \label{tab:main_table}
    \vspace{-0.4cm}
\end{figure*}

\noindent\textbf{Evaluation Setting \& Benchmark Models.}
When evaluating the performance of each model in Benchmark, we consider the most influential omnimodal models in open source and commercial APIs, including Gemini, Gemma3n, Qwen2.5-Omni and Minicpm\_o. All models are reasoned with the same set of prompts, and video samples are processed at 1fps (max 32 frames). In the open source models, all our models are run under a single H200. In order to ensure the fairness of all model tests, all models are evaluated using a uniform prompt. And the audio \& video detailed caption prompt as follow (for the rest of the questions, please refer to appendix): 

\noindent\textit{\textbf{<prompt>}You are given a short video with both audio and visual content. Write a detailed and coherent paragraph that naturally integrates all modalities. Your description should include: (1) the primary scene and background setting; (2) key characters or objects and their actions or interactions; (3) significant audio cues such as voices, background music, sound effects, and their emotional tone; (4) any on-screen text (OCR) and its role in the video context; and (5) the overall theme or purpose of the video. Ensure the output is a fluent and objective paragraph, not a bullet-point list, and captures the video's content in a human-like, narrative style.\textbf{</prompt>}}

\noindent\textbf{Metric Design.}
Our UGC‐VideoCap benchmark includes two types of questions: Open‐ended and Multiple‐Choice (see Figure~\ref{fig:sample}). For Multiple‐Choice Questions (MCQs), we follow the standard evaluation protocol and use \emph{Accuracy (ACC)}—based on exact answer matching—as the primary metric. For Open‐ended Questions, we employ GPT-4o-2024-08-06 (ensuring that neither the video nor the associated text content appears in its training set) as an automatic judge. The exact prompts used for each evaluation are provided in the Appendix. Finally, for OCR tasks within the visual‐detail category, we report the average of BLEU and ROUGE scores~\citep{papineni2002bleu,lin2004rouge}.

\noindent\textbf{Benchmark Results.}
As shown in Table~\ref{tab:main_table}, we conduct a comprehensive evaluation of full-modal models on the UGC-VideoCap benchmark, where both audio and visual inputs are provided for every question. The Gemini family demonstrates superior performance across all metrics, with \textbf{Gemini-2.5 Flash} achieving the highest overall score (\textbf{76.73}). In contrast, most open-source models show imbalanced performance between audio and visual tasks, especially on fine-grained visual details such as OCR, background transitions, and object recognition.
UGC videos pose unique multimodal challenges due to their uncontrolled environments, diverse content styles, and frequent scene changes. These factors require models to handle non-studio speech, variable sound sources, noisy visual frames, and overlapping modalities. However, current models often treat audio and visual streams independently, leading to fragmented understanding. For example, \textbf{Qwen2.5-Omni-3B} achieves an excellent voice score (\textbf{93.0}) but struggles with visual object recognition (\textbf{18.3}).
Interestingly, in final caption generation, most models (except the Gemma series) rely more heavily on visual cues than audio, suggesting limited audio integration. We believe future research should focus on enabling audio to assist visual understanding particularly in UGC settings where visual signals alone are often insufficient. Achieving balanced or audio-dominant performance could unlock better alignment with real-world multimodal understanding needs.
\section{UGC Video Captioner}
\label{sec:methods2}

\begin{figure}[t!]
    \centering
        \includegraphics[width=\textwidth]{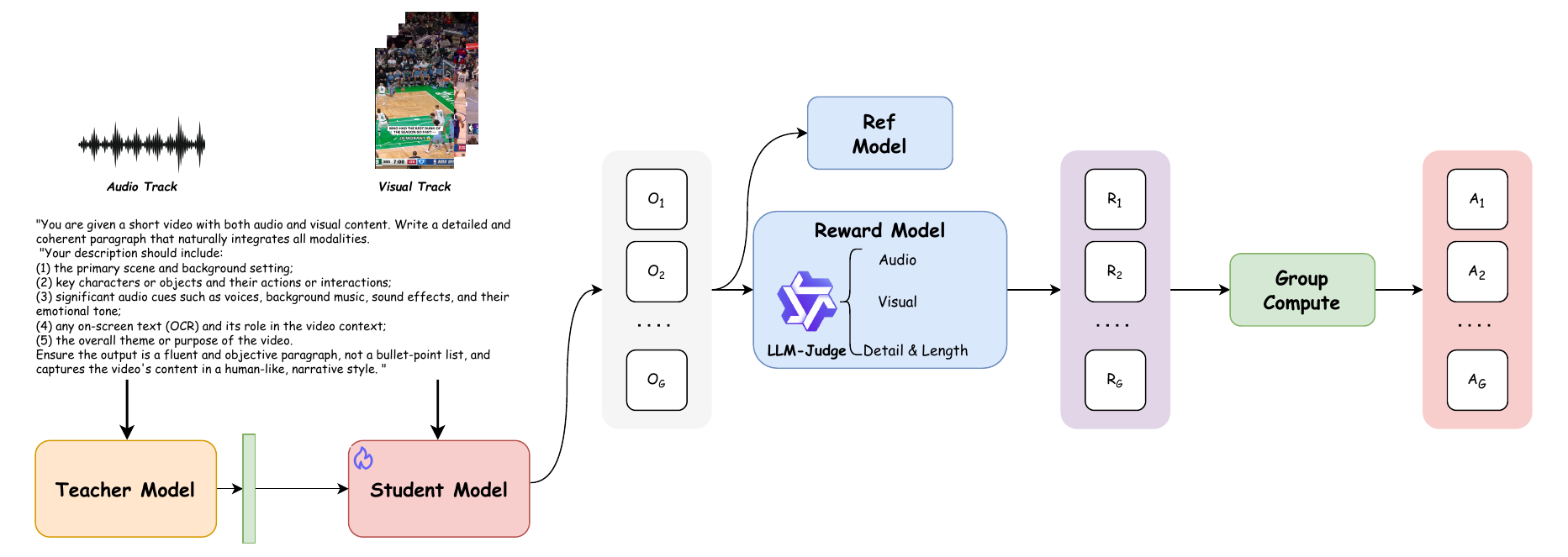}
    \caption{
    \textbf{Left:} \textit{Distillation Stage}. We use Gemini-2.5-flash to generate 20k omni video detail caption. 
    \textbf{Right:}  \textit{Reinforcement Learning Stage}. We use GRPO algorithm and new caption reward model to enhance the detail caption ability. 
    }
    \label{fig:looping_init_overview}
\end{figure}

\subsection{Distillation from Large Omni Model}
In recent years, visual multimodal large language models (MLLMs) have witnessed rapid development, with the research focus increasingly shifting from traditional large language models (LLMs) to their multimodal~\citep{liu2023visual} counterparts, particularly in the domain of video understanding, where significant advancements have been achieved. However, real-world videos, especially user-generated content (UGC), typically encompass both visual and audio modalities. While the GPT series currently lacks native support for joint audio-visual processing, Gemini has emerged as the state-of-the-art model in the omni(audio+visual) domain. To address the scarcity of high-quality audio-visual captioning datasets, we adopt a distillation-based approach using Gemini-2.5-Flash as the teacher model. We construct a dataset comprising 20,000 TikTok videos with detailed annotations and use it to train our proposed model, UGC-VideoCaptioner, which is built upon Qwen2.5-Omni-3B, for comprehensive audio-visual video captioning.

\subsection{Supervised Fine-Tuning for Cold Start}
In the absence of manually annotated data, we adopt a supervised fine-tuning strategy via \textbf{cold-start knowledge distillation}, where a smaller student model is trained to mimic the outputs of a stronger teacher model. Let $\mathcal{X} = \{x_1, x_2, \ldots, x_N\}$ denote the set of input examples. For each input $x_i$, we obtain the teacher model's response $\hat{y}_i = T(x_i)$, where $T$ denotes the teacher model. These responses act as pseudo-labels to supervise the training of the student model $S$.
The student is optimized to maximize the likelihood of reproducing the teacher's output under a teacher-forcing training paradigm. The distillation loss for a single training example is defined as:

\begin{equation}
\mathcal{L}_{\text{distill}}(x_i) = - \sum_{t=1}^{T} \log P_S\left(\hat{y}_{i,t} \mid \hat{y}_{i,<t}, x_i\right)
\end{equation}

where $\hat{y}_{i,t}$ is the $t$-th token of the teacher's generated output for input $x_i$, $\hat{y}_{i,<t}$ denotes the sequence of preceding tokens before time step $t$, and $P_S$ is the conditional token probability distribution predicted by the student model.
The overall objective is to minimize the average distillation loss over all input examples:

\begin{equation}
\mathcal{L}_{\text{total}} = \frac{1}{N} \sum_{i=1}^{N} \mathcal{L}_{\text{distill}}(x_i)
\end{equation}

This framework allows the student model to acquire domain-specific knowledge from a powerful teacher without requiring any manually labeled data, making it particularly effective in cold-start scenarios where labeled supervision is unavailable.

\subsection{Do we need GRPO for detail caption training?}
It is well known that the most direct and simple way we want to enhance a small parametric model is to distil good models. With the recent proposal of RL post-training strategies for multimodal large language models (in particular, the emergence of Deepseek's GRPO~\citep{guo2025deepseek}), it is beginning to be argued that the emergence of RL can allow for a substantial improvement in model performance through preference and reward through multiple sampling and reward function design. But without the involvement of data in the form of Chain of Thought, can GRPO further enhance the distilled model as a third stage of post-training for non-thinking models? Our chapter will explore the GRPO post-training performance of detailed captions of omni videos with increasing amounts of data.

\noindent\textbf{Revisit Group Relative Policy Optimization.}
Group Relative Policy Optimization(GRPO) is a recent policy optimization technique tailored for training large models (like language or multimodal models) with complex reasoning tasks~\citep{shao2024deepseekmath}. At its core, GRPO is similar in spirit to Proximal Policy Optimization (PPO) but introduces key differences in how the policy is updated and how the reward signal is used. Rather than relying on a learned value function (critic) to estimate expected reward, GRPO entirely foregoes the critic model. Instead, it estimates the baseline or expected reward by sampling a group of outputs from the current policy for each input and using their collective reward statistics. In other words, the model generates multiple candidate answers for a given query or state and uses their relative rewards to decide how to adjust the policy. This group-based advantage estimation removes the need to train a separate value network, simplifying the algorithm and reducing computational overhead.

So, in the second stage of training, we apply the GRPO algorithm to fine-tune UGC-VideoCaptioner with reinforcement learning. At this stage UGC-VideoCaptioner has distilled a certain amount of data through supervised finetuning, so the aim is to further improve its accuracy and detail captioning ability on omni caption task by optimizing directly for task-specific rewards. The objective of GRPO training is to adjust the model’s parameters to maximize the expected reward $\mathbb{E}[R(x,y)]$ over the distribution of video questions, using the group-based policy update scheme outlined earlier. UGC-VideoCaptioner’s initial policy for this stage $\pi_{\theta_{\text{init}}}$ is set to the fine-tuned model from stage one. We then iteratively improve the policy using GRPO updates: in each iteration, for each training query $x$, multiple answers $y_1,\dots,y_K$ are sampled and scored, and the policy is updated to prefer answers with higher scores, while maintaining closeness to $\pi_{\text{init}}$. 
\begin{equation}
J(\theta) 
= \mathbb{E}_{x}\!\biggl[
  \sum_{k=1}^K w_k \,\log \pi_{\theta}\bigl(y_k \mid x\bigr)
\biggr]
- \beta \, D_{\mathrm{KL}}\!\Bigl(
   \pi_{\theta}(\cdot \mid x)
   \,\Big\|\,
   \pi_{\mathrm{ref}}(\cdot \mid x)
\Bigr)
\end{equation}
Here $w_k$ is the weight assigned to output $y_k$ after reward normalization (for example, $w_k$ could be $\frac{\exp(\tilde{r}k/\tau)}{\sum_j \exp(\tilde{r}j/\tau)}$ for some scaling temperature $\tau$), and the second term is a KL divergence penalty. The KL term, with coefficient $\beta$, measures the divergence between the updated policy $\pi\theta$ and a reference policy $\pi{\text{ref}}$. 

Through these principles, GRPO provides a stable and efficient way to fine-tune UGC-VideoCaptioner: it pushes the model toward higher-reward outputs while maintaining coherence with its initial learned behavior.

\noindent\textbf{Reward Function for Omni Video Detail Caption.}
To further enhance the quality of omni detail video caption, we introduce a LLM-based omni reward and length-based reward to enhance the  distillation quality and regulate the length of the model's output. Specifically, this mechanism aims to enhanced the caption ability in visual, audio, detail and reduced hallucinations. So the LLM-based omni reward and length-based reward as follow. And in order to enhance the quality of distillation, we use Gemini-2.5-flash as our LLM judge model.

\noindent \textbf{\textit{The LLM-based omni reward:}}
To assess the overall quality of a predicted video description, we design a reward function based on large language model (LLM) judgment. The model is prompted to compare the predicted and ground truth descriptions across the following five key dimensions:

\begin{itemize}
  \item \textbf{scene\_background}: the primary scene and background setting,
  \item \textbf{characters\_objects}: key characters or objects and their actions or interactions,
  \item \textbf{audio\_cues}: voices, background music, sound effects, and their emotional tone,
  \item \textbf{ocr\_text}: any on-screen text (OCR) and its contextual role,
  \item \textbf{theme\_purpose}: the overall theme or purpose of the video.
\end{itemize}

The LLM provides a single integer score \( R_{\text{LLM}} \in \{1, 2, 3, 4, 5\} \) reflecting how well the predicted caption captures these five aspects. The score is determined according to predefined criteria, with lower scores indicating hallucinations, omissions, or factual inconsistencies.
\textbf{Scoring rules:} (1) If the predicted description includes hallucinated content that clearly contradicts the ground truth (e.g., non-existent objects, scenes, or sounds), the score must not exceed 2. (2) If the score appears between two levels, the lower score is always chosen to ensure conservative evaluation.

\noindent \textbf{\textit{The length-based reward:}}
In the length control, in order to enhance the effect of distillation we here let the model learn the output length distribution of Gemini-2.5-flash under different samples by controlling the length, thus letting our small-parameter model learn as much as possible the full-modal detailcaption ability of Gemini-2.5-flash.
\vspace{-0.1in}
\begin{equation}
R_l = 
\begin{cases}
0 , & \text{if } len(completions) < 0.5*len(GT) \\
0.5, & \text{if } 0.7*len(GT)> len(completions) > 0.5*len(GT) \\
1.0 & len(completions) > 0.7*len(GT)
\end{cases}
\end{equation}

\noindent\textbf{Training Setting.}
We use Qwen2.5-Omni-3B as the baseline model. In the training phase we employed distillation technique using Gemini-2.5-Flash to annotate 20,000 of TikTok as training data for SFT and RL. In the training we used 1fps (max frames set to 32 frames, and the max pixels of each frame are 100176) and max prompt length is 8192. All experiments are conducted on 8xH200-144GB GPUs.

\begin{table}[ht]
\centering
\renewcommand{\arraystretch}{1.2}
\resizebox{0.95\linewidth}{!}{
\begin{tabular}{lccccc}
\hline
Model & Training Data & \(Caption_{\ Audio}\) & \(Caption_{\ Visual}\) & \(Caption_{\ Details}\) & Average \\
\hline
Gemini-2.5-pro & - & 70.8 & 75.8 & 74.8 & 73.78 \\
Gemini-2.5-flash & - & \textbf{74.2} & \textbf{78.8} & \textbf{77.2} & \textbf{76.73} \\
Qwen2.5-Omni-3B & - & 48.2 & 55.6 & 52.6 & 52.18 \\
\rowcolor{blue!10}Qwen2.5-Omni-3B & 1k SFT & 61.4 & 58.4 & 57.0 & 58.96 (+6.78) \\
\rowcolor{blue!10}Qwen2.5-Omni-3B & 10k SFT & 63.2 & 58.4 & 58.0 & 59.87 (+7.69) \\
\rowcolor{blue!10}Qwen2.5-Omni-3B & 20k SFT & 64.0 & 59.2 & 58.4 & 60.50 (+8.32) \\
\rowcolor{green!10}UGC-VideoCaptioner-3B-zero & 1k RL & 53.0 & 57.8 & 55.4 & 55.40\textbf{ (+3.22)} \\
\rowcolor{green!10}UGC-VideoCaptioner-3B & 1k SFT + 1k RL & 62.4 & 59.4 & 58.2 & 60.01\textbf{ (+7.83)} \\
\hline
\end{tabular}}
\caption{Performance comparison on audio-visual captioning. Gemini models serve as upper bounds. Reinforcement learning (RL) and supervised fine-tuning (SFT) on small datasets substantially improve caption quality.}
\vspace{-5mm}
\label{caption_table}
\end{table}

\noindent\textbf{Training Results.}
Based on the above theoretical algorithm design and training setup, we conducted a detailed experimental demonstration of the traditional sft distillation method and our two-stage distillation method under the 3B model. As shown in Table~\ref{caption_table}, we conducted supervised fine-tuning (SFT) with datasets of three different sizes: 1k, 10k, and 20k. The results indicate that the 3B model quickly approaches a performance bottleneck under SFT. Specifically, doubling the data from 10k to 20k yields only a marginal improvement of 0.63 points, suggesting diminishing returns and raising concerns about the efficiency of further data scaling for small models.
To address this, we introduced a new distillation strategy combining SFT with reinforcement learning (SFT+RL), alongside a redesigned reward function. Remarkably, this approach, using only 1k samples for SFT and an additional 1k for RL, not only outperformed the 10k SFT baseline but also closely approached the performance of the 20k SFT setting.
While RL clearly enhances model performance, we further investigated whether it could replace SFT entirely. Experimental comparisons between 1k RL and 1k SFT show that SFT still provides a stronger performance gain. This suggests that RL serves best as a complementary stage—refining a model that has already undergone supervised distillation—rather than as a standalone substitute.
These findings highlight that, in the context of detailed caption modeling, large-scale distillation datasets are not strictly necessary to reach the performance ceiling of small models. Instead, a carefully designed two-stage training pipeline (SFT followed by RL) can achieve competitive performance with significantly less data.

\section{Conclusion and Future Work}

In this work, we presented \textbf{UGC-VideoCap}, a large-scale benchmark and training framework tailored for evaluating and developing multimodal language models on detailed audio-visual understanding in UGC video scenarios. By introducing structured QA tasks and annotations for audio, visual, and final captions, UGC-VideoCap addresses the limitations of prior benchmarks that neglect the audio modality. Our proposed model, \textbf{UGC-VideoCaptioner-3B}, leverages a two-stage training strategy—first distilling knowledge from a stronger teacher (Gemini-2.5 Flash), then applying reinforcement learning with GRPO and LLM-based rewards—to significantly improve caption quality while remaining lightweight and efficient.
Looking ahead, we identify several promising directions. First, integrating automatic audio event detection and voice diarization into the annotation pipeline could further enrich the benchmark. Second, incorporating multilingual audio and text capabilities would enable broader applicability to diverse social media contexts. Lastly, we encourage the community to explore adaptive inference strategies and modality-aware attention mechanisms to better handle the noisy and heterogeneous nature of UGC content. We hope UGC-VideoCap can serve as a catalyst for advancing real-world omnimodal video understanding.


\clearpage
{
\bibliography{googledeepmind}
}

\clearpage
\counterwithin{figure}{section}
\counterwithin{table}{section}
\appendix



\end{document}